\documentclass[runningheads]{llncs}
\usepackage[utf8]{inputenc}
\usepackage{graphicx}
\usepackage{amssymb,amsmath,mathtools}
\usepackage{enumitem}
\usepackage[pdftex, pdfborder={0 0 0}]{hyperref}
\usepackage[textsize=small]{todonotes}
\usepackage{pgfplots}
\usepackage{tikz}
\usepackage{csquotes}
\usepackage{siunitx}
\usepackage{marvosym}
\usepackage{algorithm2e}
\usepackage{multirow}
\usepackage{subcaption}

\newcommand{\cosim}{\operatorname{sim}} %
\newcommand{\ve}[1]{\mathbf{#1}} %
\newcommand{\tsum}{\sum\nolimits}
\newcommand{\argmax}{\ensuremath{\operatorname{arg\,max}}}

\newcommand{\norm}[1]{\left\lVert#1\right\rVert}

\newcommand{\sprod}[2]{\left<#1,#2\mathstrut\right>}

\newcommand{\reffig}[1]{Fig.~\ref{#1}}
\newcommand{\reftab}[1]{Table~\ref{#1}}
\newcommand{\refeqn}[1]{Eq.~\refeq{#1}}

\begin{document}

\title{Accelerating Spherical $k$-Means%
\thanks{Part of the work on this paper has been supported by Deutsche Forschungsgemeinschaft (DFG),
project number 124020371,
within the Collaborative Research Center SFB 876
``Providing Information by Resource-Constrained Analysis'', project A2 
}}
\author{Erich~Schubert\Letter\orcidID{0000-0001-9143-4880}
\and
Andreas~Lang\orcidID{0000-0003-3212-5548}
\and
Gloria~Feher\orcidID{0000-0002-0859-2042} 
}
\titlerunning{Accelerating Spherical $k$-Means}
\authorrunning{E. Schubert et al.}
\institute{TU Dortmund University, Dortmund, Germany 
\email{\{erich.schubert,andreas.lang,gloria.feher\}@tu-dortmund.de}
}
\maketitle

\begin{abstract}
Spherical $k$-means is a widely used clustering algorithm for sparse
and high-dimensional data such as document vectors.
While several improvements and accelerations have been introduced for
the original $k$-means algorithm, not all easily translate to the
spherical variant: Many acceleration techniques,
such as the algorithms of Elkan and Hamerly, rely on the triangle
inequality of Euclidean distances.
However, spherical $k$-means uses Cosine similarities instead of
distances for computational efficiency. In this paper, we incorporate
the Elkan and Hamerly accelerations to the spherical $k$-means algorithm
working directly with the Cosines instead of Euclidean distances to
obtain a substantial speedup and evaluate
these spherical accelerations on real data.

\end{abstract}

\section{Introduction}

Clustering textual data is an important task in data science with applications
in areas like information retrieval, topic modeling, and knowledge organization.
Spherical $k$-means~\cite{DBLP:journals/ml/DhillonM01} is a widely used adaptation
of the $k$-means clustering algorithm to high-dimensional sparse data,
such as document vectors where Cosine similarity is a popular choice.
While it is generally used for clustering documents, it has also been applied
to medical images~\cite{Arfiani/2019,DBLP:conf/midp/MoriyaRNONOM18},
multivariate species occurrence data~\cite{Hill/species/2013},
and plant leaf images~\cite{Alamoudi/plant/2020}. 
Because of its importance, several improvements and extensions have been suggested.
Many optimizations improve the initialization of $k$-means cluster centers,
such as $k$-means++~\cite{DBLP:conf/soda/ArthurV07} and
$k$-means$||$~\cite{DBLP:conf/icml/BachemL017,DBLP:journals/pvldb/BahmaniMVKV12},
some of which have also been adapted to spherical $k$-means~%
\cite{DBLP:conf/mdai/EndoM15,DBLP:conf/acml/PratapDND18,DBLP:journals/jgo/LiXZZ20}.

A key area of optimizations is focussed on the iterative optimization phase of $k$-means.
The standard algorithm computes the distance of every point to every cluster
in each iteration. Many of these computations are not necessary if cluster centers have
not moved much, and hence a lot of research has been on how to avoid computing distances.
The central work in this domain is the algorithm of
Elkan~\cite{DBLP:conf/icml/Elkan03}, which is the base for many other variants such
as Hamerly's algorithm~\cite{DBLP:conf/sdm/Hamerly10}, but also recently
the Exponion algorithm~\cite{DBLP:conf/icml/NewlingF16}, the Shallot algorithm~\cite{DBLP:conf/ida/Borgelt20},
and the variants of Yu et al.~\cite{DBLP:conf/sisap/YuCC20},
all of which rely on the Euclidean triangle inequality
to avoid distance computations.

This paper studies how to adapt such acceleration techniques to spherical $k$-means, thus providing a more efficient approach for clustering text documents. %

\section{Foundations}

Cosine similarity (which we will simply denote using $\cosim$ in the following)
is commonly defined as the Cosine of the angle $\theta$ between two vectors $\ve{x}$ and $\ve{y}$:
$$
\cosim(\ve{x},\ve{y}):=
\operatorname{sim}_{\text{Cosine}}(\ve{x},\ve{y}) :=
\frac{\sprod{\ve{x}}{\ve{y}}}{\norm{\ve{x}}_2\cdot \norm{\ve{y}}_2}
=
\frac{\sum_i x_iy_i}{\sqrt{\sum_i x_i^2}\cdot \sqrt{\sum_i y_i^2}}
=\cos \theta
$$
In the following, we will only consider vectors normalized to unit length,
i.e., with Euclidean norm $\norm{\ve{x}}_2{=}1$. It is trivial to see that on
such vectors, the Cosine similarity is simply the dot product.
Consider the Euclidean distance of two \emph{normalized} vectors $\ve{x}$ and $\ve{y}$,
and expand using the binomial equations, we obtain:
\begin{align}
d_{\text{Euclidean}}(\ve{x},\ve{y}) &:=
\sqrt{\rule[-5pt]{0pt}{14pt}\smash{\tsum_i (x_i - y_i)^2}}
=
\sqrt{\rule[-5pt]{0pt}{14pt}\smash{\tsum_i (x_i^2 + y_i^2 - 2x_iy_i)}}
\\
&=
\sqrt{\norm{\ve{x}}^2+\norm{\ve{y}}^2 -2 \sprod{\ve{x}}{\ve{y}}}
=
\sqrt{2 - 2 \cdot \cosim(\ve{x},\ve{y})}
\label{eqn:euclid}
\end{align}
where the last step relies on the vectors being normalized. Hence we have an
extremely close relationship between Cosine similarity and squared Euclidean distance
on normalized vectors: $\cosim(\ve{x},\ve{y}) = 1 - \tfrac12 d_{\text{Euclidean}}^2(\ve{x},\ve{y})$.

$k$-means minimizes the \emph{squared} Euclidean distances of points to their cluster centers
and hence can be used to maximize Cosine similarities.
Because the total variance of a data set is constant,
by minimizing the within-cluster squared deviations,
$k$-means also maximizes the between-cluster squared deviations.
By adapting this to Cosine, we obtain clusters where objects in the same cluster have to be more similar,
while objects in different clusters are less similar.

Dhillon and Modha~\cite{DBLP:journals/ml/DhillonM01}
popularized this idea as ``spherical $k$-means'' for clustering text documents
and exploited exactly the above relationship
between the squared Euclidean distance and Cosine similarity. Only a tiny modification
of the standard $k$-means algorithm is necessary to obtain the desired results:
the arithmetic mean of a cluster usually does not have unit Euclidean length.
Hence, after recomputing the cluster mean, we normalize it accordingly.
This constrains the clustering to split the data at great circles (i.e., hyperplanes
through the origin), rather than arbitrary Voronoi cells as with regular $k$-means.

On text data, computing the Cosine similarity is more efficient than computing Euclidean
distance because of sparsity: rather than storing the vectors as a long array of values,
most of which are zeros, only the non-zero values can be encoded as pairs $(i,v)$ of an index $i$
and a value $v$, and stored and kept in sorted order.
The dot product of two such vectors can then be efficiently computed by a \textit{merge} operation,
where only those indexes $i$ need to be considered that are contained in both vectors, because in
$\sprod{\ve{x}}{\ve{y}}{=}\sum_i x_i y_i$ only those terms matter where both $x_i$ and $y_i$
are not zero.
A merge is most efficient if both vectors are sparse, but even the dot product of a sparse and
a dense vector is often much faster than that of two dense vectors.
While we can also compute Euclidean distance this way (using Eq.~\refeq{eqn:euclid}), this
computation is prone to the numerical problem called ``catastrophic cancellation'' for small
distances that can be problematic in clustering (see, e.g., \cite{DBLP:conf/ssdbm/SchubertG18,DBLP:conf/sisap/LangS20}).
Hence, working with Cosines directly is preferable.

Instead of recomputing the distances to all cluster centers, the idea of
algorithms such as Elkan's is to keep an upper bound
on the distance to the nearest cluster, and one or more lower bounds on the distances
to the other centers. Let $d_n$ be the distance to the nearest center, $d_n{\leq} u$ an upper bound,
$d_s$ the distance to the second nearest, and $l{\leq} d_s$ a lower bound. If we have $u{\leq} l$, then
the nearest cluster must still be the same since $d_n{\leq} u {\leq} l {\leq} d_s$.
Updating the distance bounds uses the triangle inequality: if the nearest center $\mu_n$ has moved
to $\mu_n'$, then $d(x,\mu_n')\leq d(x,\mu_n) + d(\mu_n,\mu_n')$; and we hence
can obtain an upper bound $u$ by adding every movement of a cluster center to the previous distance.
Lower bounds are obtained similarly: starting with the initial distance as the lower bound,
we subtract the distance the other center has moved to obtain a provable new lower bound.
While Elkan stored a lower bound for each cluster (which needs $O(N{\cdot}k)$ memory),
Hamerly~\cite{DBLP:conf/sdm/Hamerly10} reduced the memory usage by using just one lower bound
to the second nearest cluster, updated by the largest distance moved.
Additional pruning rules involve the pairwise distances of centers~\cite{DBLP:conf/icml/Elkan03},
annuli around centers~\cite{DBLP:conf/icml/NewlingF16}, and the relative movement of centers~\cite{DBLP:conf/sisap/YuCC20}.

In the following, we describe how such accelerations can be applied to spherical $k$-means, i.e., for Cosine similarity and high-dimensional data.

\section{Pruning with Cosine Similarity}

Many acceleration techniques %
rely on the triangle inequality of the (non-squared) Euclidean distance.
Hence, we can adapt these methods by computing Euclidean distances from our
Cosine similarities using $d_{\text{Euclidean}}(\ve{x},\ve{y}){=}\sqrt{2-2\cdot\cosim(\ve{x},\ve{y})}$,
but we would like to avoid this because of
(i)~the square root takes 10--50 CPU cycles (depending on the exact CPU, precision, and input value)
and (ii)~the risk of numerical instability because of catastrophic cancellation.
Hence we develop techniques that directly use similarities instead of distances,
yet allow a similar pruning to these (very successful) acceleration techniques of regular $k$-means.

The arc length (i.e., the angle $\theta$ itself, rather than the Cosine of the angle) satisfies
the triangle inequality and hence we could use
\begin{align}
\cosim(\ve{x},\ve{y}) &\geq \cos(\arccos(\cosim(\ve{x},\ve{z})) + \arccos(\cosim(\ve{z},\ve{y})))
\enskip,
\label{eq:tri-acos}
\end{align}
but unforunately the trigonometric functions in here are even more expensive (60--100 CPU cycles each).
Schubert~\cite{conf/sisap/Schubert21} recently proposed the following more efficient reformulations
avoiding the expensive trigonometric functions:
\begin{align}
\cosim(\ve{x},\ve{y}) &\geq \cosim(\ve{x},\ve{z})\cdot \cosim(\ve{z},\ve{y})
- \sqrt{(1{-}\cosim(\ve{x},\ve{z})^2)\cdot (1{-}\cosim(\ve{z},\ve{y})^2)}
\label{eq:tri-cosine}
\\
\cosim(\ve{x},\ve{y}) &\leq \cosim(\ve{x},\ve{z})\cdot \cosim(\ve{z},\ve{y})
+ \sqrt{(1{-}\cosim(\ve{x},\ve{z})^2)\cdot (1{-}\cosim(\ve{z},\ve{y})^2)}
\label{eq:tri-cosine2}
\end{align}
In this paper, we explain how to integrate these triangle inequalities into spherical
$k$-means, and discuss an easily overlooked pitfall therein.

\section{Upper and Lower Bounds}

In the following, we orient ourselves on the very concise presentation
and notation of Hamerly~\cite{DBLP:conf/sdm/Hamerly10} as well as
Newling and Fleuret~\cite{DBLP:conf/icml/NewlingF16},
except that we swap the names of $u$ and $l$, because switching from
distance to similarity requires us to swap the roles of upper and lower bounds.
We will assume that all points are normalized to unit length, and hence
$\cosim(\ve{x},\ve{y})=\sprod{\ve{x}}{\ve{y}}=\ve{x}^T\!{\cdot}\ve{y}$.

The algorithms we discuss will employ upper and lower bounds for the
similarities of each sample $x(i)$ to the cluster centers $c(j)$.
$l(i)$ is a lower bound for the similarity to the current cluster $a(i)$,
$u(i,j)$ are upper bounds on the similarity of each point to each cluster center,
respectively $u(i)$ is an upper bound on the similarity to all other cluster centers
($u(i,j)$ and $u(i)$ are used in different variants, not at the same time).
These bounds are maintained to satisfy:
\begin{alignat*}{3}
l(i) &\leq \sprod{x(i)}{c(a(i))}
&\qquad
u(i,j) &\geq \sprod{x(i)}{c(j)}
&\qquad
u(i) &\geq \max_{j\neq a(i)}\sprod{x(i)}{c(j)}
\end{alignat*}
The central idea of all the discussed variants is that if we have
$l(i){\geq} u(i,j)$, then $\sprod{x(i)}{c(a(i))} {\geq} l(i) {\geq} u(i,j) {\geq} \sprod{x(i)}{c(j)}$
implies that the current cluster assignment of object $x(i)$ is optimal,
and we do not need to recompute the similarities. %

The bounds $l(i)$ and $u(i,j)$, can be maintained using above triangle inequality
if we know how much the cluster centers $c(j)$ moved from their previous location $c'(j)$.
Let $p(j){:=}\sprod{c(j)}{c'(j)}$ denote this similarity.
Based on the triangle inequalities \refeqn{eq:tri-cosine} and \refeqn{eq:tri-cosine2}, we obtain the following bound update
equations:
\begin{align}
l(i) &\leftarrow l(i)\cdot p(a(i))-\sqrt{(1-l(i)^2)\cdot(1-p(a(i))^2)}
\label{eqn:l-bound} 
\\
u(i,j) &\leftarrow u(i,j)\cdot p(j)+\sqrt{(1-u(i,j)^2)\cdot(1-p(j)^2)}  
\label{eqn:u-elkan} 
\end{align}

\section{Accelerated Spherical $k$-Means}

The algorithms discussed here all follow
the outline of the standard $k$-means algorithm of alternating optimization.
During initialization, all data samples $x(i)$ are normalized to have length $\norm{x(i)}{=}1$.
In the first step, all objects are reassigned to the nearest cluster, in the
second step, the cluster center is optimized. However, we switch the notation
from distance to similarity.
Let the variable $a(i)$ denote the current cluster assignment of sample $x(i)$,
and denote the current cluster centers using $c(j)$, the two steps can be written as:
\begin{alignat*}{2}
a(i) &\leftarrow \argmax_j \sprod{x(i)}{c(j)}
&\qquad i&\in 1..N
\\
c(j) &\leftarrow
\frac{\sum_{i\mid a(i)=j} x(i)}{\norm{\rule[-4pt]{0pt}{12pt}\smash{\sum_{i\mid a(i)=j} x(i)}}}
& j &\in 1..k
\end{alignat*}
When computing $a(i)$ we maximize the Cosine similarity instead of the squared Euclidean
distance in regular $k$-means.
For $c(j)$, note that the denominator is different here, as we want to have $\norm{c(j)}{=}1$ for all $j$.
We hence do not need to compute the arithmetic mean, but we can scale the sum directly to length~1.

There are several optimizations we can do for the baseline
algorithm that make a difference: %
(i)~By normalizing the vectors, we do not have to take the vector lengths of $x(i)$ into account
when updating $c(j)$, and by also normalizing the $c(j)$ we can use the dot product when computing $a(i)$.
(ii)~Both the dot product as well as the sum operation when computing $c(j)$ can be optimized for sparse data.
(iii)~Instead of recomputing $c(j)$ each time, it is better to store the sums before normalization
and update them when a cluster assignment changes.

\subsection{Spherical Simplified Elkan's Algorithm}

As the name suggests, this algorithm is a simplified version of Elkan's approach,
introduced by Newling and Fleuret~\cite{DBLP:conf/icml/NewlingF16}. As it uses a
subset of the pruning rules, we introduce it before Elkan's full algorithm.
Both are presented directly in the adaptation for spherical $k$-means.

Simplified Elkan uses the test $u(i,j) {\leq} l(i)$ to skip computing
the similarity between $x(i)$ and $c(j)$ when it is not necessary.
If this test fails, $l(i){\leftarrow}\sprod{x(i)}{c(a(i))}$ is updated first
(as the current assignment is clearly the best guess), and only if
the condition still is violated, $u(i,j){\leftarrow}\sprod{x(i)}{c(j)}$
is computed next, and the point is reassigned if necessary
(updating $l(i)$ and $a(i)$ then).

\subsection{Spherical Elkan's Algorithm}

Elkan's algorithm~\cite{DBLP:conf/icml/Elkan03} uses additional
tests based on the pairwise distance of centers,
respectively pairwise cluster similarities here.
The idea is that cluster centers are supposedly
well separated, whereas points are close to their nearest cluster,
and we can use half the distance between two centers as a threshold.
We simplify the computation of half of the angle per
$\cos(\frac{1}{2}\arccos(x)){=}\sqrt{(x{+}1)/2}$.
Let $cc(i,j){:=}\sqrt{(\sprod{c(i)}{c(j)}{+}1)/2}$ be this lower bound
($cc$ for center-center bounds, as in \cite{DBLP:conf/icml/NewlingF16}).
Let $s(i){:=}\max_{j\neq i}cc(i,j)$ denote the maximum such
bound for each~$i$.

\emph{Suppose} that $cc(a(i),j)\leq l(i)$ and $l(i)\geq 0$, then $\sprod{c(i)}{c(j)} \leq 2l(i)^2 - 1$.
We can then use \refeqn{eq:tri-cosine2} to bound the distance to another
cluster $c(j)\neq c(a(i))$ per
\begin{align*}
\sprod{x(i)}{c(j)}
&\leq \sprod{x(i)}{c(a(i))}\cdot\sprod{c(a(i))}{c(j)}
\notag
\\&\phantom{={}}
+ \sqrt{\rule[-2pt]{0pt}{10pt}\smash{(1{-}\smash{\sprod{x(i)}{c(a(i))}}^2)\cdot (1{-}\smash{\sprod{c(a(i))}{c(j)}}^2)}}
\\
&\leq l(i)(2l(i)^2 - 1) + \sqrt{(1{-}l(i)^2)\cdot (1{-}(2l(i)^2 - 1)^2)}
\\
&= 2l(i)^3-l(i) + \sqrt{(1{-}l(i)^2)\cdot 4l(i)^2(1{-}l(i)^2)}
\\
&= 2l(i)^3-l(i) + 2l(i)(1{-}l(i)^2)
= l(i)
\quad,
\end{align*}
and hence do not have to consider other cluster centers $c(j)$
\emph{if} $cc(a(i),j)\leq l(i)$.
Because $s(i)$ is the maximum of these values, we can skip iterating over the means
if $s(i)\leq l(i)$ altogether.
While these additional tests are fairly cheap to compute, they were found to not
always be effective by Newling and Fleuret~\cite{DBLP:conf/icml/NewlingF16}
(who, hence, suggested the simplified variant discussed in the previous section).

For spherical $k$-means clustering, these bounds may not be very
effective because of the high dimensionality. Using these bounds adds
$k\cdot(k-1)/2=O(k^2)$ similarity computations to each iteration.
Furthermore, the necessary computations can become more expensive because
the centers are best stored using dense vectors because
(i)~we aggregate many vectors into each
center, and only attributes zero in all of the assigned vectors will be zero
in the resulting center, i.e., the sparsity decreases often to the
point where a dense representation is more compact, and (ii)~the efficient sparse data
structures we use for the $x(i)$ are not well suited for adding and removing attributes.
We could aggregate into a dense vector and convert it to a sparse
representation when normalizing the center, but the resulting vectors will
still often be too dense to be efficient. %

\subsection{Spherical Hamerly's Algorithm}
Where Elkan's algorithm used one upper bound for each cluster,
Hamerly~\cite{DBLP:conf/sdm/Hamerly10} only uses a single bound
for all clusters.
This does not only saves memory (for large $k$,
memory consumption of Elkan's algorithm can be an issue)
but updating the $N{\cdot} k$ bounds each iteration even if the clusters
change only very little takes a considerable amount of time.
Hamerly's idea is to make a worst-case assumption, where we use the
distance to the second nearest center as the initial bound, and update it based on
the largest cluster movement (of all clusters, except the one currently assigned to).
Because of this, the bound will become loose much faster, and hence we
need to recompute more often (and then we need to recompute the distances
to all clusters). Because of this, it is hard to predict which algorithm
works better, we are trading reduced memory and fewer bound updates against
additional distance computations.
Nevertheless, many later works have confirmed that it is often favorable to
only keep one bound.

At first, adapting Hamerly to Cosine similarity appears to be straightforward.
To obtain the lowest upper bound per object $u(i){\leq} \min_{j\neq a(i)} u(i,j)$,
we would compute the smallest similarity of a cluster center to its previous location
(as well as the second smallest, in case the point is currently assigned to that center),
then use \refeqn{eqn:u-elkan} with $p'(i){:=}\min_{j\neq i} p(j)$
(which is either the smallest or the second smallest $p(j)$).
Most of the time this is fine, but there is a subtly hidden catch here because
of the underlying non-monotone trigonometric functions.

Recall the update equation \eqref{eqn:u-elkan}, rewritten
to $u(i)$ instead of $u(i,j)$ already:
\begin{align*}
u(i) &\leftarrow u(i)\cdot p(j)+\sqrt{(1-u(i)^2)\cdot(1-p(j)^2)}
\end{align*}
This equation is not necessarily minimized by the smallest $p(j)$, because
of the square root term. For large $u(i)$ (e.g., 1), the result will be determined
by the first term, and a smaller $p(j)$ is what is needed. But for small $u(i)$
(e.g., 0), the second term becomes influential, and a larger $p(j)$ causes
a smaller bound. This is because we are working with the Cosines $\cos\theta$,
not the angles $\theta$ themselves.
Unfortunately, this depends on the previous value of $u(i)$,
and we probably cannot use just one $p(j)$ for all points.

One option would be to use both the minimum $p'(i){:=}\min_{j\neq i} p(j)$
and the maximum $p''(i){:=}\max_{j\neq i} p(j)$
to update the bound with:
\begin{align}
\eqref{eqn:u-elkan} &\leq
u(i)\cdot p''(a(i))+\sqrt{(1-u(i)^2)\cdot(1-p'(a(i))^2)}
\enskip.
\label{eqn:u-hamerly-1}
\intertext{
Because $p''(j)\rightarrow 1$ as the algorithm converges,
we may omit this term entirely:
}
\eqref{eqn:u-hamerly-1} &\leq
u(i)+\sqrt{(1-u(i)^2)\cdot(1-p'(a(i))^2)}
\label{eqn:u-hamerly-2}
\end{align}
This has almost identical pruning power once $p''(j)$ becomes large enough in later iterations.
As we can precompute $(1-p'(j))$ for all $j$, this is quite efficient.
We cannot rule out that a tighter and computationally efficient bound exists.

If the condition $l(i){\geq} u(i)$ is violated, first $l(i)$ is made tight again,
and if it still is violated, all remaining similarities are computed to update $u(i)$,
or to potentially obtain a new cluster assignment (updating $a(i)$, $l(i)$, and $u(i)$).

\subsection{Spherical Simplified Hamerly's Algorithm}

Hamerly's algorithm contains a bounds test similar to Elkan's algorithm,
but using only the distance of each center to its nearest neighbor center
instead of keeping all pairwise center distances, i.e., only the threshold
$s(i){:=}\max_{j\neq i} cc(i,j)$ to prune objects with $l(i){\geq} s(a(i))$.
We also consider a ``simplified'' variant of Hamerly's algorithm in our experiments
with this bound check removed for the same reasons as discussed with Elkan's algorithm.

\subsection{Further $k$-Means Variants}
An obvious candidate to extend this work is Yin-Yang $k$-means~\cite{DBLP:conf/icml/DingZSMM15},
which groups the cluster centers and uses one bound for each group.
This is a compromise between Elkan's and Hamerly's approaches,
encompassing both as extreme cases ($k$ groups respectively one group).
The results of this paper will trivially transfer to this method.
The Annulus algorithm~\cite{doi:10.1007/978-3-319-09259-1_2} additionally uses the distance
from the origin for pruning. As all our data is normalized to unit length,
this approach clearly will not help for spherical $k$-means.
The Exponion~\cite{DBLP:conf/icml/NewlingF16} and Shallot~\cite{DBLP:conf/ida/Borgelt20} algorithms
transfer this idea to using pairwise distances of cluster centers,
where our considerations may be applicable again.

\subsection{Spherical $k$-means++}

We experiment with the canonical adaptation of $k$-means++, using the analogy with
squared Euclidean distance. The first sample is chosen uniformly at random,
the remaining instances are sampled proportional to $1-\max_c\sprod{x(i)}{c}$
which is proportional to the squared Euclidean distance used by $k$-means++.
This can be done in $O(nk)$ by caching the previous maximum, and the scalar
product is efficient for two sparse vectors.
Endo and Miyamoto~\cite{DBLP:conf/mdai/EndoM15} prove theoretical guarantees
for a slight modification of spherical $k$-means using the
dissimilarity of $\alpha-\sprod{\ve{x}}{\ve{y}}$ with $\alpha\geq\tfrac32$
to make it metric, and hence sample proportionally to $1-\max_c\sprod{x(i)}{c}$.
Pratap et al.~\cite{DBLP:conf/acml/PratapDND18} use the same trick to
apply the AFK-MC\textsuperscript{2} algorithm~\cite{DBLP:conf/nips/BachemLH016}
to spherical $k$-means-clustering.

\section{Experiments}

We implemented our algorithms in the Java framework ELKI~\cite{DBLP:journals/corr/abs-1902-03616},
which already contained a large collection of $k$-means variants. By keeping the
implementation differences to a minimum, we try to make the benchmark experiments
more reliable (c.f., \cite{DBLP:journals/kais/KriegelSZ17}), but the caveats
of Java just-in-time compilation remain.

\begin{table}[tb]\centering
\caption{Data sets used in the experiments.}
\label{tab:datasets}
\setlength{\tabcolsep}{5pt}
\begin{tabular}{l|rrr}
Data set & Rows & Columns & Non-zero
\\\hline
DBLP Author-Conference & 1842986 & 5236 & 0.056\%
\\
DBLP Conference-Author & 5236 & 1842986 & 0.056\%
\\
DBLP Author-Venue & 2722762 & 7192 & 0.099\%
\\
Simpsons Wiki & 10126 & 12941 & 0.463\%
\\
20 Newsgroups & 11314 & 101631 & 0.096\%
\\
Reuters RCV-1 & 804414 & 47236 & 0.160\%
\end{tabular}
\end{table}

As our method is designed for sparse and high-dimensional data sets,
we focus on textual and graph data as input.
From the Digital Bibliography \& Library Project (DBLP, \cite{DBLP:conf/spire/Ley02})
we extracted graphs that connect authors and conferences.
As this includes many authors with just a single
paper, the data set is very sparse. We can either use the authors as samples
and the conferences as columns or transposed. But because we use TF-IDF weighting
afterward the semantics will be different. 
Spherical $k$-means clustering has been used successfully for
community detection on such data sets (although we have to choose the number of communities as a parameter).
If we also include journals, the data set becomes both larger and denser.
A second data set was obtained from the Simpsons Fandom Wiki,\footnote{\url{https://simpsons.fandom.com/wiki/Simpsons_Wiki}}
from which we extracted the text of around 10000 articles. The text was tokenized and lemmatized,
stop words were removed as well as infrequent tokens (reducing the dimensionality from 42124 to 12941,
and increasing the density of non-zero values from 0.153\% to 0.463\%).
This data set is more typical of a smaller domain-specific text corpus.
The 20 Newsgroups data set is a classic, popularized by the textbook of Tom Mitchell.
We use a version available via scikit-learn, with headers, footers, and quotes removed and
vectorized using the default settings (i.e., TF-IDF weighting). This is much more sparse
than the Simpsons wiki because of the poor input data quality (including Base64-encoded attachments).
After removing stop words and rare words as above, the density would have been 0.317\%,
but we opted for the default scikit-learn version instead.
Reuters RCV-1~\cite{DBLP:journals/jmlr/LewisYRL04} is another classic text categorization benchmark,
with a density between the Simpsons and the 20news data.

\begin{figure}[tb]\centering
\includegraphics[width=0.7\linewidth]{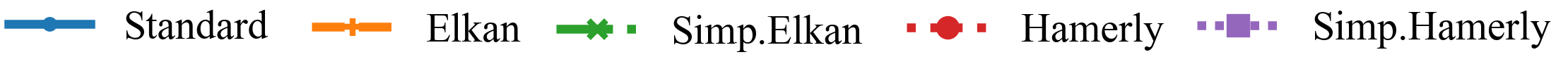}\\
\begin{subfigure}{.48\linewidth}\centering
\includegraphics[width=\linewidth]{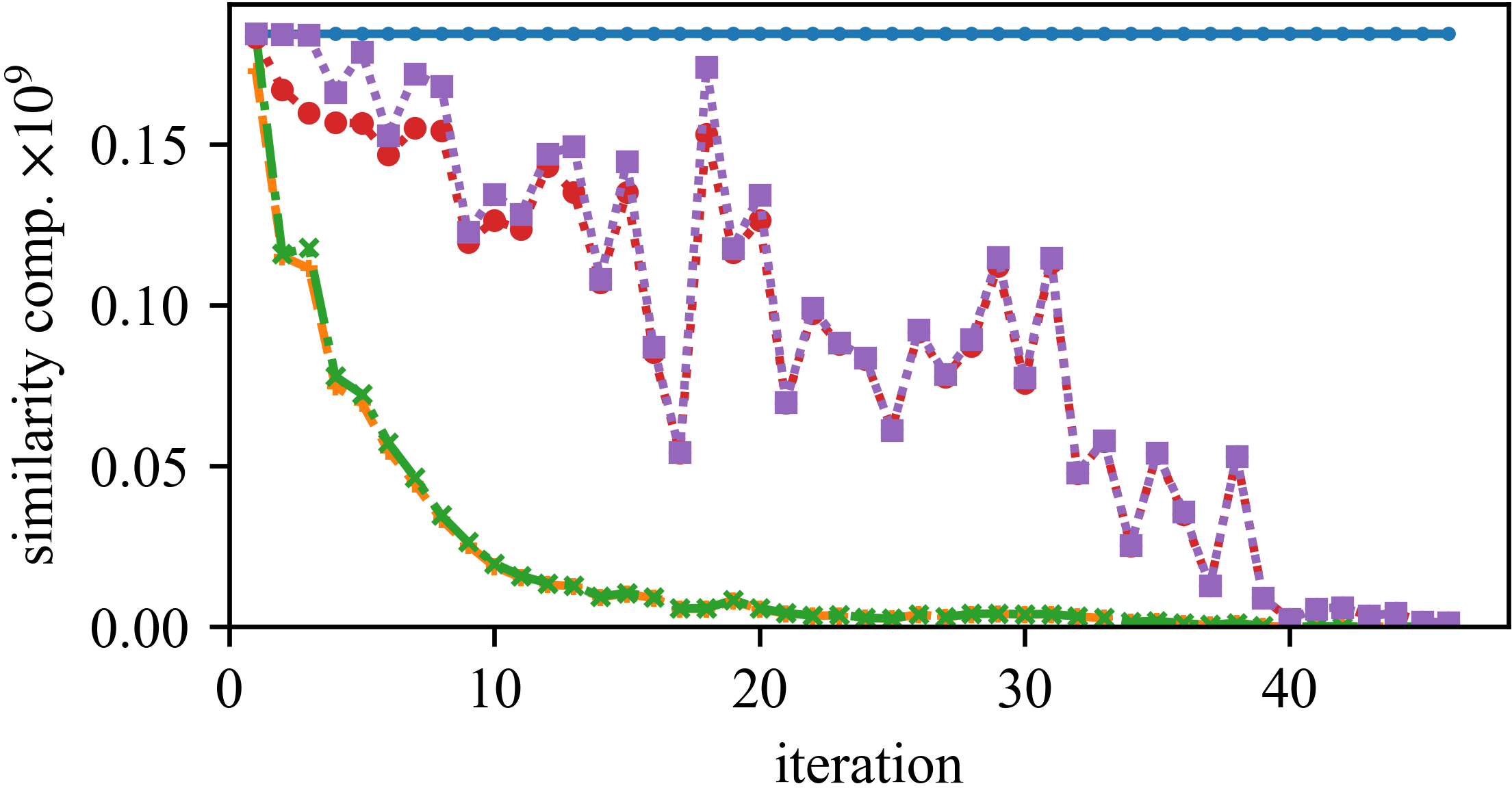}
\caption{Similarity computations per iteration.}
\label{fig:sample-sims}
\end{subfigure}%
\hfill%
\begin{subfigure}{.48\linewidth}\centering
\includegraphics[width=\linewidth]{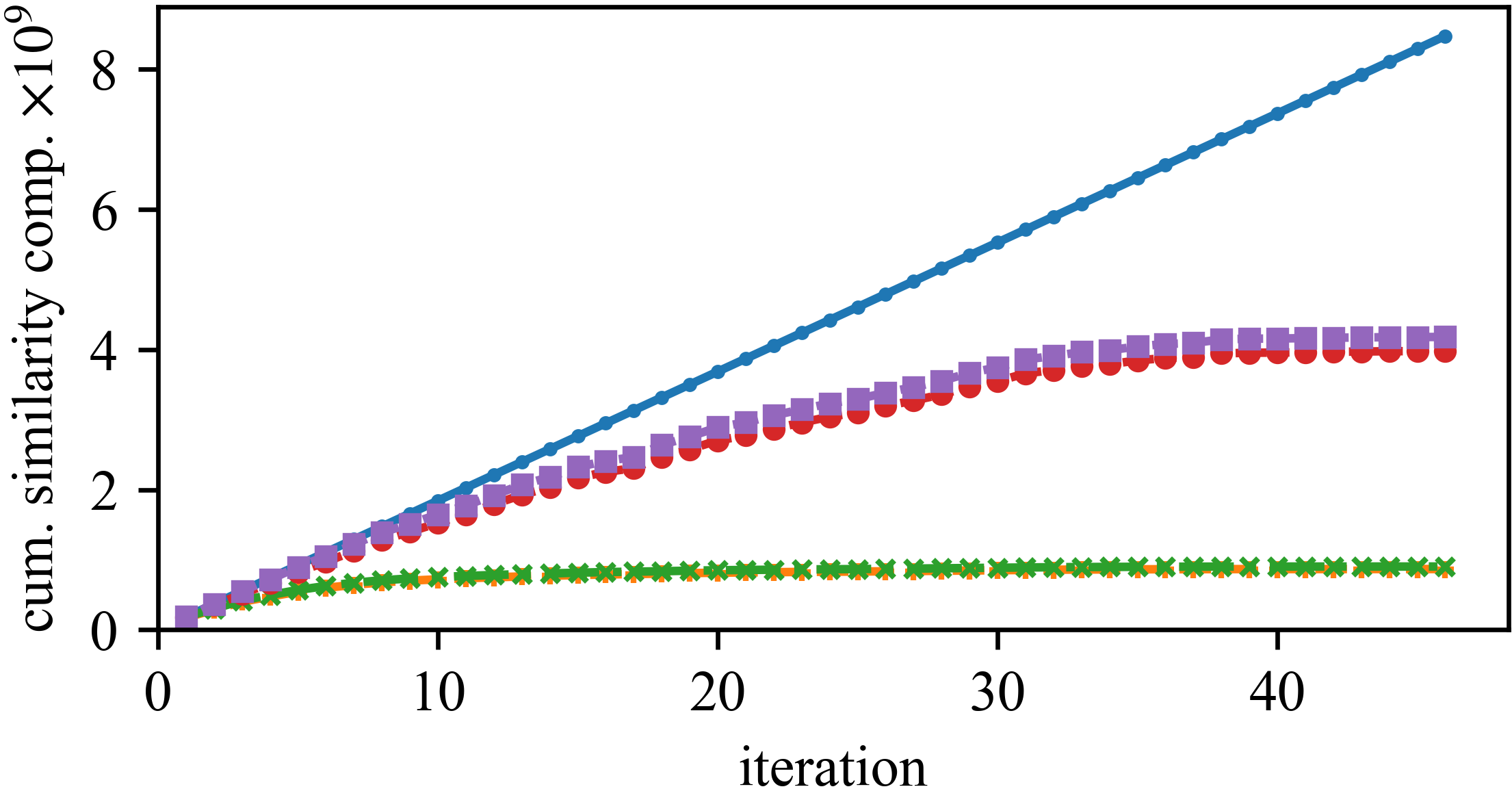}
\caption{Number of similarity computations.}
\label{fig:sample-csims}
\end{subfigure}%
\\
\begin{subfigure}{.48\linewidth}\centering
\includegraphics[width=\linewidth]{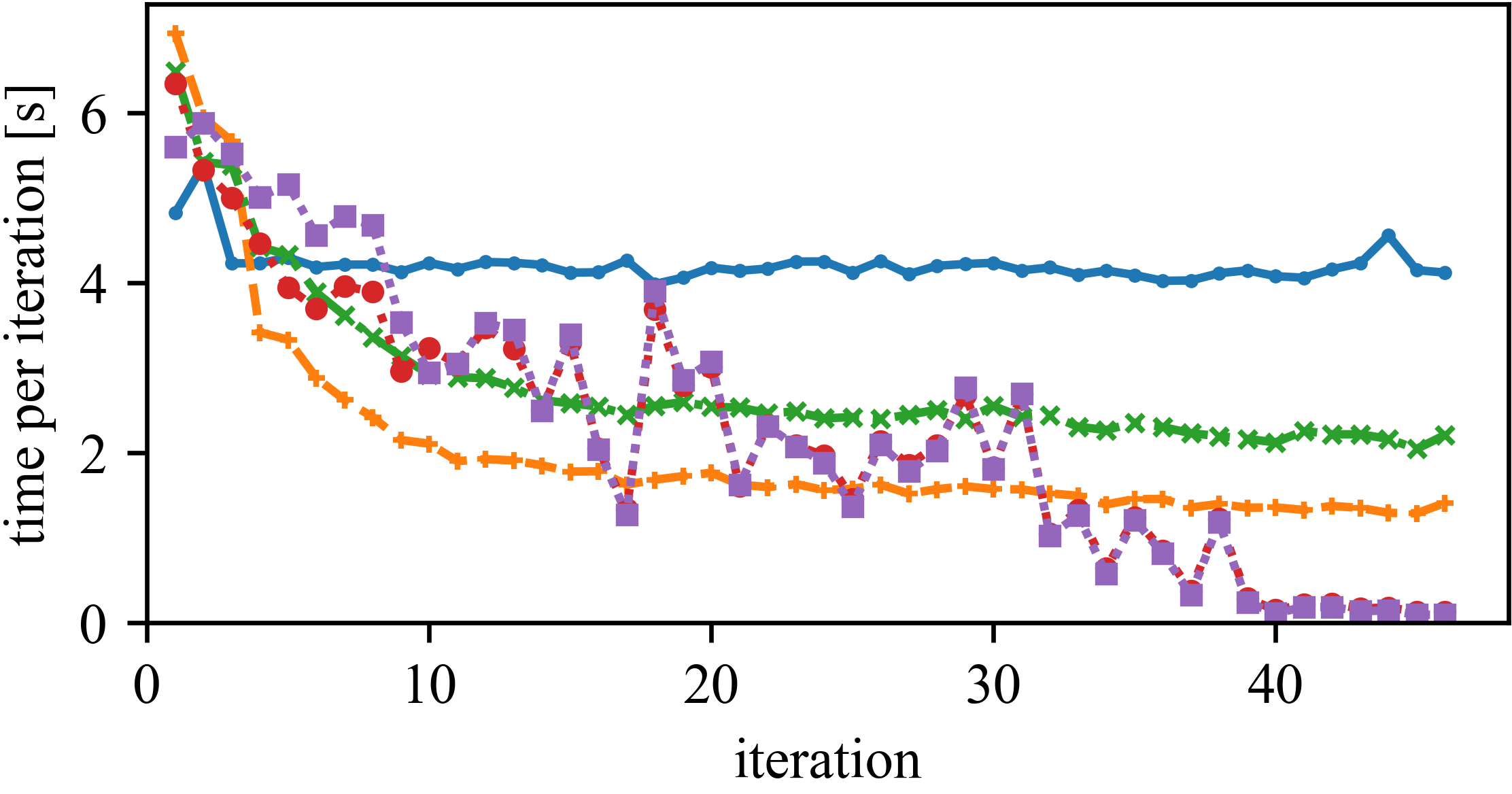}
\caption{Run time per iteration.}
\label{fig:sample-time}
\end{subfigure}
\hfill%
\begin{subfigure}{.48\linewidth}\centering
\includegraphics[width=\linewidth]{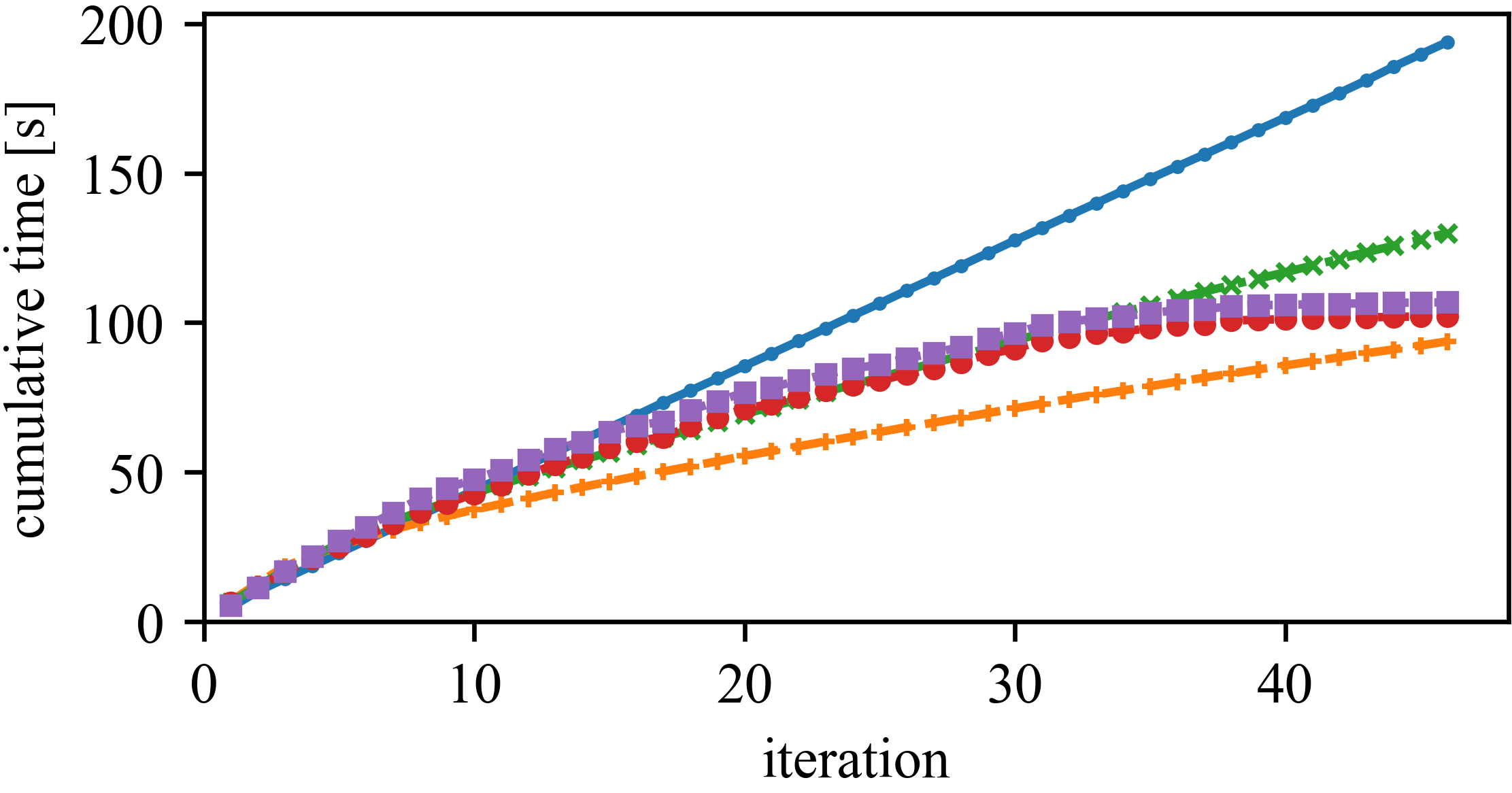}
\caption{Cumulative run time.}
\label{fig:sample-ctime}
\end{subfigure}
\caption{Distance computations and resulting run time for one initialization.}
\label{fig:sample}
\end{figure}

We first discuss the algorithms on a single data set, with a single random seed,
averaged over 10 re-runs, to observe some characteristic behavior.
The reason that we do not average over different random initializations is that
we want to observe individual iterations of the algorithms, which depend
on the initialization.
\reffig{fig:sample} shows the results on the DBLP author-conference data set
with a large $k{=}100$.
Considering only similarity computations (\reffig{fig:sample-sims} and \reffig{fig:sample-csims}),
both Elkan and Simplified Elkan shine (as expected) and use the fewest computations as
they have the tightest bounds. There is next to no difference among the two concerning the
number of computations, but considering the run time (\reffig{fig:sample-time} and \reffig{fig:sample-ctime})
the simplified variant is much worse. Perhaps unexpectedly, this can be explained by the high $k$.
The additional pruning rule of the full algorithm allows skipping the loop over all clusters $k$,
which would otherwise each have to be checked against their bounds.
The behavior of the Hamerly variants is much more chaotic because it only depends on the cluster center that
changed most. Because of this, Hamerly computes many more similarities than Elkan until the last few iterations.
Nevertheless, its total run time is initially similar to that of Simplified Elkan, and after around 30 iterations
its run time per iteration (c.f., \reffig{fig:sample-time}) becomes even lower than the full Elkan algorithm's.
These savings arise once clusters do not change much anymore because only 2 bounds need to be updated instead of $k{+}1$ bounds per iteration.
For $k{=}10$ (not shown in the figures), both Hamerly variants outperform Elkan, while for $k{=}1000$ even Simplified Elkan clearly
outperforms both Hamerly variants.
Note that we used random sampling as initialization.
If we had known the optimal initial cluster centers, all methods would have converged instantly.

\begin{table}[bt!]\centering
\caption{Relative change in the objective function compared to the random initialization (lower is better).}
\label{tab:init}
\begin{tabular}{ll|rrrrrr}
Data set                        & Initialization & k=2              & k=10             & k=20             & k=50             & k=100            & k=200            \\ \hline
\multirow{5}{*}{Simpsons Wiki}     & Uniform                                                          & 0.00\%           & 0.00\%           & 0.00\%           & 0.00\%           & 0.00\%           & 0.00\%           \\
    & $k$-means++ $\alpha {=} 1$                                                         & -0.27\%          & \textbf{-0.16\%} & \textbf{-0.24\%} & -0.07\%          & -0.18\%          & -0.07\%          \\
    & $k$-means++ $\alpha {=} 1.5$                                                         & -0.16\%          & -0.13\%          & -0.17\%          & -0.01\%          & -0.18\%          & \textbf{-0.09\%} \\ 
    & AFK-MC\textsuperscript{2}~ $\alpha {=} 1$                                                         & \textbf{-0.44\%} & 0.12\%           & -0.15\%          & \textbf{-0.15\%} & \textbf{-0.24\%} & -0.08\%          \\
    & AFK-MC\textsuperscript{2}~ $\alpha {=} 1.5$                                                         & -0.31\%          & 0.21\%           & 0.09\%           & 0.09\%           & -0.05\%          & -0.02\%          \\\hline
\multirow{5}{*}{\shortstack[l]{DBLP\\Author-Conf.}}    & Uniform                                                          & 0.00\%           & 0.00\%           & 0.00\%           & 0.00\%           & 0.00\%           & 0.00\%           \\
    & $k$-means++ $\alpha {=} 1$                                                        & \textbf{-0.11\%} & 0.12\%           & -0.07\%          & 0.27\%           & 0.14\%           & \textbf{-1.67\%} \\
    & $k$-means++ $\alpha {=} 1.5$                                                         & -0.03\%          & 0.11\%           & 0.33\%           & 0.68\%           & 0.53\%           & -0.74\%          \\ 
    & AFK-MC\textsuperscript{2}~ $\alpha {=} 1$                                                         & -0.01\%          & \textbf{-0.06\%} & \textbf{-0.87\%} & \textbf{-0.47\%} & -0.48\%          & -1.03\%          \\
    & AFK-MC\textsuperscript{2}~ $\alpha {=} 1.5$                                                         & -0.03\%          & 0.34\%           & -0.32\%          & 0.09\%           & \textbf{-0.56\%} & -1.10\%          \\\hline
\multirow{5}{*}{\shortstack[l]{DBLP\\Author-Venue}} 
    & Uniform                                                          & 0.00\%           & 0.00\%           & 0.00\%           & 0.00\%           & 0.00\%           & 0.00\%           \\
    & $k$-means++ $\alpha {=} 1$                                                        & -0.13\%          & 0.09\%           & -0.12\%          & 0.13\%           & \textbf{-0.74\%} & \textbf{-1.70\%} \\
    & $k$-means++ $\alpha {=} 1.5$                                                         & -0.01\%          & 0.18\%           & 0.00\%           & 0.23\%           & 0.39\%           & -0.17\%          \\
    & AFK-MC\textsuperscript{2}~ $\alpha {=} 1$                                                        & -0.17\%          & 0.10\%           & -0.05\%          & 0.47\%           & -0.20\%          & -0.68\%          \\
    & AFK-MC\textsuperscript{2}~ $\alpha {=} 1.5$                                                         & \textbf{-0.19\%} & \textbf{-0.33\%} & \textbf{-0.68\%} & \textbf{-0.04\%} & -0.50\%          & -1.41\%          \\ \hline
\multirow{5}{*}{\shortstack[l]{DBLP\\Conf.-Author}}   
    & Uniform                                                          & \textbf{0.00\%}  & \textbf{0.00\%}  & \textbf{0.00\%}  & 0.00\%           & 0.00\%           & 0.00\%           \\
    & $k$-means++ $\alpha {=} 1$                                                        & 0.01\%           & 0.04\%           & 0.05\%           & -0.02\%          & -0.13\%          & -0.09\%          \\
    & $k$-means++ $\alpha {=} 1.5$                                                         & \textbf{0.00\%}  & 0.11\%           & 0.08\%           & \textbf{-0.15\%} & -0.18\%          & \textbf{-0.13\%} \\ 
    & AFK-MC\textsuperscript{2}~ $\alpha {=} 1$                                                         & 0.04\%           & \textbf{0.00\%}  & 0.05\%           & -0.10\%          & \textbf{-0.19\%} & -0.02\%          \\
    & AFK-MC\textsuperscript{2}~ $\alpha {=} 1.5$                                                         & 0.04\%           & \textbf{0.00\%}  & 0.06\%           & -0.12\%          & -0.15\%          & -0.06\%          \\ \hline
\multirow{5}{*}{20 Newsgroups}           
    & Uniform                                                          & \textbf{0.00\%}  & \textbf{0.00\%}  & \textbf{0.00\%}  & \textbf{0.00\%}  & \textbf{0.00\%}  & \textbf{0.00\%}  \\
    & $k$-means++ $\alpha {=} 1$                                                        & 0.38\%           & 0.52\%           & 0.78\%           & 1.83\%           & 4.09\%           & 7.34\%           \\
    & $k$-means++ $\alpha {=} 1.5$                                                         & 0.72\%           & 0.93\%           & 0.89\%           & 2.39\%           & 4.65\%           & 7.87\%           \\ 
    & AFK-MC\textsuperscript{2}~ $\alpha {=} 1$                                                         & 0.24\%           & 0.31\%           & 0.31\%           & 0.41\%           & 0.11\%           & 0.23\%           \\
    & AFK-MC\textsuperscript{2}~ $\alpha {=} 1.5$                                                         & 0.37\%           & 0.17\%           & 0.26\%           & 0.30\%           & 0.08\%           & 0.23\%           \\ \hline
\multirow{5}{*}{RCV-1}            
    & Uniform                                                          & 0.00\%           & 0.00\%           & 0.00\%           & 0.00\%           & 0.00\%           & \textbf{0.00\%}  \\
    & $k$-means++ $\alpha {=} 1$                                                        & 0.13\%           & \textbf{-0.11\%} & 0.08\%           & \textbf{-0.25\%} & \textbf{-0.17\%} & 0.06\%           \\
    & $k$-means++ $\alpha {=} 1.5$                                                         & -0.03\%          & 0.21\%           & 0.53\%           & 0.44\%           & 0.04\%           & 0.16\%         \\
    & AFK-MC\textsuperscript{2}~ $\alpha {=} 1$                                                         & \textbf{-0.24\%} & -0.01\%          & \textbf{-0.03\%} & 0.39\%           & 0.05\%           & 0.24\%           \\
    & AFK-MC\textsuperscript{2}~ $\alpha {=} 1.5$                                                         & 0.13\%           & 0.07\%           & \textbf{-0.03\%} & 0.22\%           & -0.09\%          & 0.15\%           
\end{tabular}
\end{table}

Next, we compare the quality and run time of the initialization methods. 
\reftab{tab:init} shows the difference in the sum of variances, averaged over 10 random seeds, compared to the uniform random initialization.
It shows that the quality difference of the converged solutions between all initialization methods is small
except for the 20-news data set where $k$-means++ performs up to 8\% worse. Supposedly, because
this data set contains anomalies.
AFK-MC\textsuperscript{2}~\cite{DBLP:conf/nips/BachemLH016} with $\alpha {=} 1$ finds the best initialization most of the time.
While $k$-means++ with $\alpha {=} 1.5$ does not quite reach same the quality, it performs generally better than uniform random.
With $\alpha {=} 1.5$, both initialization methods are worse and more often than not are below the quality of the random uniform initialization.
The run time behavior is similar on all data sets.
The uniform initialization is nearly instantaneous, while the $k$means++ and AFK-MC\textsuperscript{2} initialization generally stay below the time needed for one iteration. They only have a small impact on the overall run time.
Usually, $\alpha{=}1$ seems to work better than $\alpha{=}1.5$, where the first is the standard Cosine similarity,
while the latter was used in the proofs to obtain a metric.

\begin{table}[bt!]\centering
\caption{Run times of all $k$-means variants in milliseconds.}
\label{tab:cluster_runtime}
\begin{tabular}{ll|rrrrrr}
Data set                        & Algorithm & k=2           & k=10           & k=20           & k=50            & k=100           & k=200           \\ \hline
\multirow{5}{*}{Simpsons Wiki}   & Standard   & 166           & 457            & 845            & 1,646            & 3,015            & 10,047           \\   
    & Elkan  & 161           & 352            & 532            & 1,198            & 2,657            & 8,247            \\
    & Simp.Elkan & \textbf{145}  & \textbf{312}   & \textbf{436}   & \textbf{800}    & \textbf{1,230}   & \textbf{3,100}   \\
    & Hamerly  & 171           & 434            & 732            & 1,860            & 3,976            & 14,386           \\
    & Simp.Hamerly & 166           & 421            & 657            & 1,450            & 2,471            & 9,858            \\
    \hline
\multirow{5}{*}{\shortstack[l]{DBLP\\Author-Conf.}}    & Standard   & 32,228         & 29,865          & 24,687          & 42,229           & 50,851           & 80,553           \\
    & Elkan  & 5,675          & 9,650           & 12,366          & 39,652           & 54,901           & 82,732           \\
    & Simp.Elkan & 5,732          & 10,841          & 15,514          & 44,991           & 66,731           & 105,905          \\
    & Hamerly  & \textbf{4,220} & 7,072           & 9,834           & 19,988           & \textbf{30,846}  & 55,687           \\
    & Simp.Hamerly & 4,285          & \textbf{7,002}  & \textbf{9,810}  & \textbf{19,690}  & 31,589           & \textbf{55,250}  \\
     \hline
\multirow{5}{*}{\shortstack[l]{DBLP\\Author-Venue}} & Standard   & 33,359         & 46,328          & 50,596          & 70,772           & 80,218           & 199,230          \\ 
    & Elkan  & 5,730          & 14,593          & 22,733          & 59,725           & 84,011           & 165,756          \\
    & Simp.Elkan & 5,986          & 16,822          & 27,200          & 68,577           & 103,678          & 209,835          \\
    & Hamerly  & 4,321          & 11,410          & 18,056          & 33,881           & \textbf{51,242}  & 125,066          \\
    & Simp.Hamerly & \textbf{4,188} & \textbf{11,096} & \textbf{17,799} & \textbf{33,017}  & 52,593           & \textbf{123,931} \\
    \hline
\multirow{5}{*}{\shortstack[l]{DBLP\\Conf.-Author}}   & Standard   & 1,149          & 6,017           & 9,672           & 20,908           & 33,973           & 61,680           \\
    & Elkan  & 943           & 5,549           & 11,907          & 41,078           & 108,028          & 32,103          \\
    & Simp.Elkan & \textbf{894}  & \textbf{4,018}  & \textbf{6,184}  & \textbf{10,998}  & \textbf{16,435}  & \textbf{29,093}  \\
    & Hamerly  & 944           & 6,840           & 14,760          & 50,282           & 125,513          & 347,668          \\
    & Simp.Hamerly & 944           & 5,347           & 9,158           & 20,115           & 32,640           & 55,421           \\
     \hline
\multirow{5}{*}{20 Newsgroups}          & Standard   & \textbf{101}  & \textbf{234}   & 1,223           & 6,755            & 16,394           & 38,131           \\
     & Elkan  & 118           & 269            & 498            & 6,683            & 19,917           & 83,407           \\
    & Simp.Elkan & 118           & 251            & \textbf{342}   & \textbf{1,876}   & \textbf{3,915}   & \textbf{7,891}   \\
    & Hamerly  & 111           & 272            & 536            & 9,542            & 28,005           & 109,204          \\
    & Simp.Hamerly & 121           & 266            & 443            & 5,298            & 12,653           & 29,915           \\
     \hline
\multirow{5}{*}{RCV-1}            & Standard   & 24,569         & 153,170         & 224,939         & 917,894          & 2,669,733         & 6,064,203  \\
    & Elkan  & 7,639          & 38,199          & 47,963          & \textbf{115,275} & \textbf{260,924} & 547,110          \\
    & Simp.Elkan & 8,825          & \textbf{41,162} & \textbf{50,161} & 123,428          & 263,728          & \textbf{474,800} \\
    & Hamerly  & \textbf{5,424} & 49,041          & 80,793          & 325,433          & 1,132,352         & 3,181,667         \\
    & Simp.Hamerly & 5,498          & 47,977          & 81,593          & 320,677          & 1,144,947         & 3,266,234
\end{tabular}
\end{table}

\begin{figure}[tb!]\centering
\includegraphics[width=0.7\linewidth]{exp/convergence/legend.png}\\
\begin{subfigure}{.48\linewidth}\centering
\includegraphics[width=\linewidth]{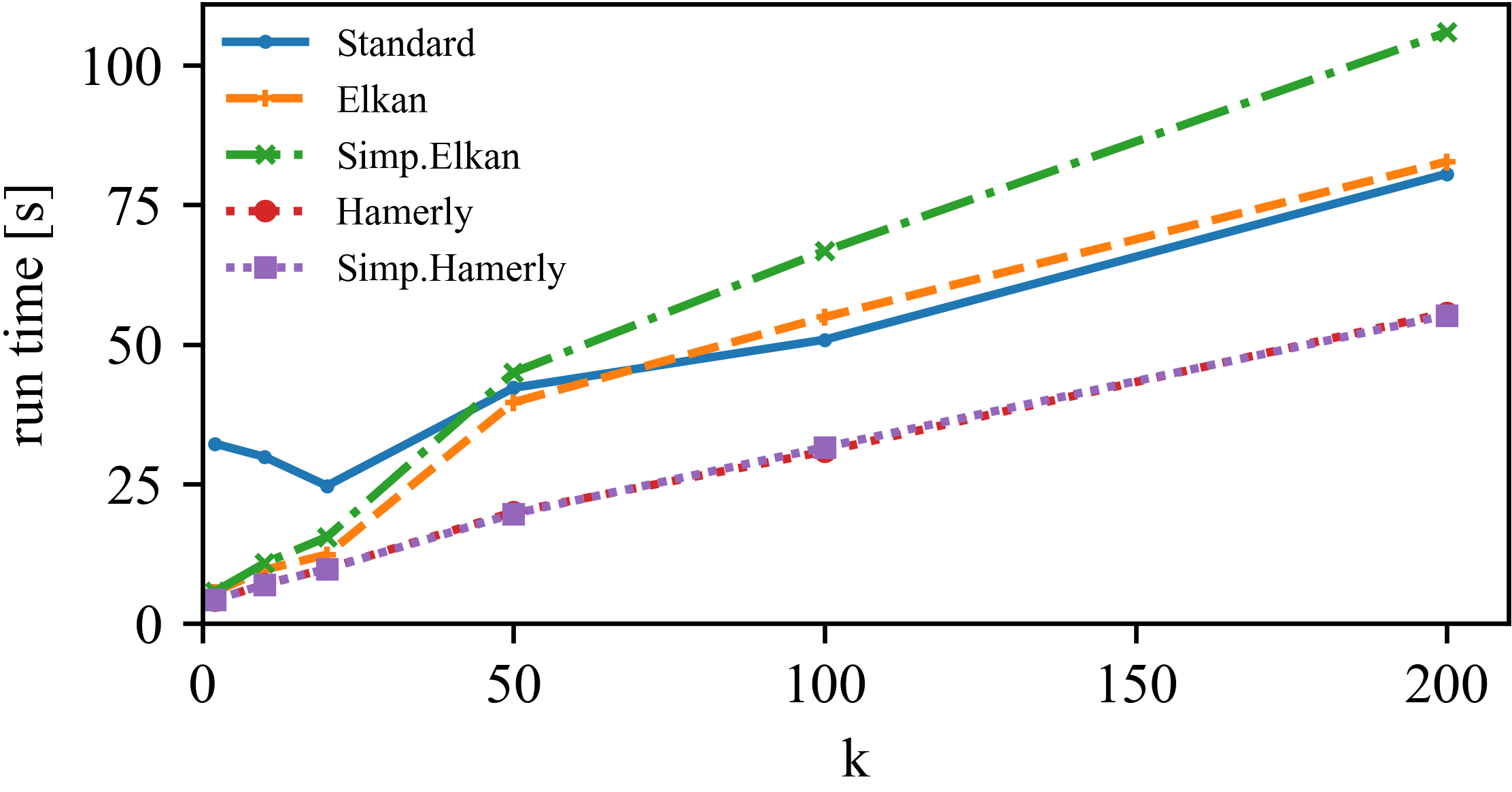}
\caption{Run time of the different algorithms on Authors-Conf. (higher $N$, lower $d$).}
\end{subfigure}
\hfill
\begin{subfigure}{.48\linewidth}\centering
\includegraphics[width=\linewidth]{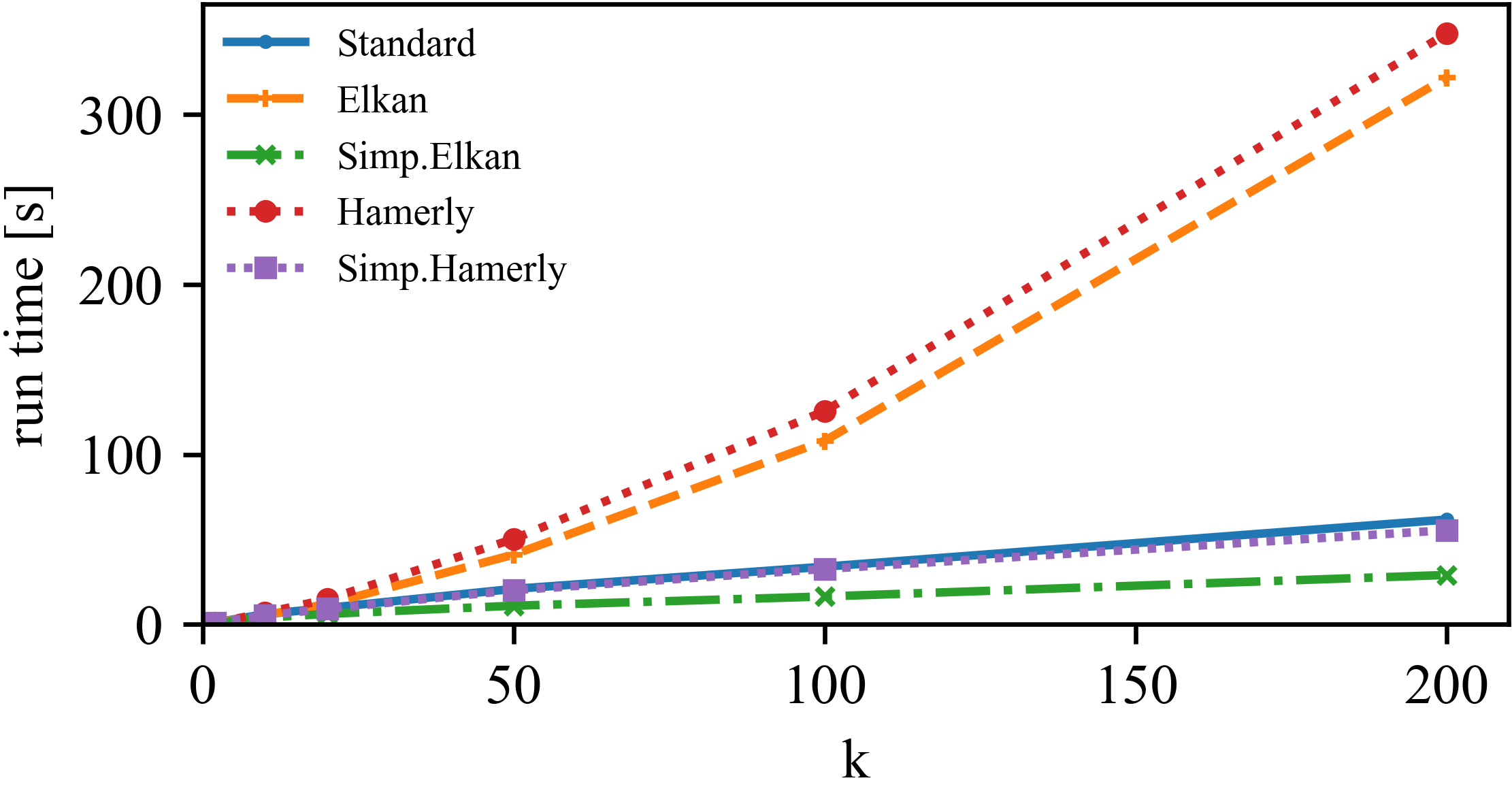}
\caption{Run time of the different algorithms on Conf.-Authors (lower $N$, higher $d$).}
\end{subfigure}
\caption{Run times of the different algorithms on the DBLP author-conference data set, and its transpose, with very different characteristics.}
\label{fig:dim_time}
\end{figure}

At last, we discuss the achieved improvements in run time for the accelerated spherical $k$-means algorithms.
As with the other experiments, each one was repeated 10 times with various random seeds.
\reftab{tab:cluster_runtime} shows that for most data sets the simplified Elkan algorithm is the fastest, but there are several interesting observations to be made.
On the Author-Conference data set, which has the most rows of all data sets but also the lowest number of columns, the normal Elkan and both Hamerly variants are faster.
Interestingly, this changes when we increase the number of columns in relation to the number of rows by transposing the data (before applying TF-IDF), shown in \reffig{fig:dim_time}. Here, the normal Elkan and Hamerly variants increase drastically in their run time when $k$ increases.
This effect originates in the increasing cost of calculating the distances between cluster centers for the additional pruning step. By transposing the data (to cluster conferences, not authors),
we increased the dimensionality by $350\times$, while at the same time reducing the number of instances
by the same factor. Computing the pairwise cluster distances now became a substantial effort.
This shows that there is no ``one size fits all'', but the best $k$-means variant needs to be chosen
depending on data characteristics such as dimensionality and the number of instances.
While Simplified Hamerly is among the best methods in both situations, it barely outperforms the standard
algorithm on the latter data set. Supposedly because of the very high dimensionality, its pruning power is
rather limited.
While the spherical Hamerly and Elkan implementations can be faster than the standard spherical $k$-means algorithms,
this depends on the data, and with an unfavorable data set they can be much worse.
The simplified version of spherical Hamerly seems to be a reasonable default choice, but for small $k$,
it may often be outperformed by the Elkan variants.
On the well-known RCV-1 data set, speedups of over $10\times$ are achievable for $k{\geq}100$.
It may be a bit disappointing that there is no ``winner'' solution, but data sets simply may have very
different characteristics. Possibly some simple heuristics can be identified to automatically choose
an appropriate alternative based on empirical thresholds (which need to be determined for a particular implementation
and hence are outside the scope of a scientific paper) on the data dimensionality and data set size.
In many cases, the limiting factor may be the memory usage and bandwidth for the Elkan variants.
Consider the DBLP authors-conference data set with $k=100$, the bounds used by Elkan with
double precision require 2~GB of RAM for the bounds alone, and have to be read and written each iteration.
The Hamerly variants only add an overhead of 44~MB. The Yin-Yang variant which we did not yet implement
allows choosing the number of bounds to use, and hence make better use of the available RAM.

\vfill
\pagebreak
\section{Conclusions}

In this article, we use the triangle inequality for Cosine similarity of Schubert~\cite{conf/sisap/Schubert21},
to accelerate spherical $k$-means clustering by avoiding unnecessary similarity computations.
We were able to adapt the well-known algorithms of Elkan and Hamerly (along with some simplified variants)
to work with similarities rather than distances throughout the algorithm. This is desirable because
the similarities are more efficient to compute, and the trigonometric bounds are tighter
than the Euclidean bounds~\cite{conf/sisap/Schubert21} (with the first corresponding to the arc length,
the latter to the chord length).

We integrated the new triangle inequality into Elkan's and Hamerly's algorithm as two prominent
and popular choices, but acknowledge there exist further improved algorithms such as the
Yin-Yang, Exponion, and Shallot algorithms that deserve attention in future work.
The purpose of this paper is to demonstrate that we can perform pruning directly on the
Cosine similarities now and that it can speed up the algorithm run times considerably
(we observed speedups of over $10\times$ for the well-known RCV-1 data set).

For further speedups, the new technique can also be combined with improved initialization methods
from literature. There exists a synergy between some initialization methods that we are not yet
exploiting in our implementation, where, e.g., the $k$-means++ initialization can pre-initialize the
bounds used here, and will then allow pruning computations already in the first iteration of the main
algorithm.

We hope that this article spurs new research on further accelerating spherical $k$-means clustering
using the triangle inequality, similar to Euclidean $k$-means.

\subsubsection*{Acknowledgments}
A simpler approach of adapting Hamerly's and Elkan's algorithms for spherical $k$-means clustering
still using Euclidean distances and not the Cosine triangle inequalities
was explored by our student, Alexander Voß, in his bachelor thesis.

\bibliographystyle{splncs04}
\bibliography{literature}
\end{document}